\documentclass[preprint,12pt]{elsarticle}

\usepackage[utf8]{inputenc}
\usepackage[T1]{fontenc}
\usepackage{amsmath,amssymb}
\usepackage{booktabs}
\usepackage{graphicx}
\usepackage{tabularx}
\usepackage{multirow}
\usepackage{enumitem}
\usepackage{xcolor}
\usepackage[hidelinks]{hyperref}
\hypersetup{
  pdftitle={When Can Fraud Operations Authorize Automation? A Decision-Support Framework for Fresh Audit Evidence and Review Workload},
  pdfauthor={Jie Deng},
  pdfsubject={Preprint; not peer reviewed}
}
\usepackage[margin=1in]{geometry}
\usepackage{setspace}
\graphicspath{{figures/}}
\journal{Preprint}
\biboptions{numbers,sort&compress}

\newcommand{\fcac}{\textsc{FCAC}}
\newcommand{\risk}{\mathcal{R}}

\begin{document}

\begin{frontmatter}

\title{When Can Fraud Operations Authorize Automation? A Decision-Support Framework for Fresh Audit Evidence and Review Workload}

\author{Jie Deng}
\address{Tongji University, Shanghai, China}
\ead{dengjie.work@foxmail.com}

\begin{abstract}
Fraud operations must allocate events among automatic approval, analyst review, and automatic blocking even though the labels needed to evaluate these actions are selective and delayed. Predictive scores order cases, but they do not show whether the evidence is current and representative enough to delegate an action to the model. We develop freshness-constrained audit capacity (\fcac), a decision-support framework that treats automation as an authorization decision constrained by action risk, evidence freshness, and shared review capacity. It evaluates candidate action regions from mature randomized audits and a prespecified temporal allowance. Supported regions are automated; unsupported regions remain in review. The resulting decision record reports evidence age, audit demand, total review workload, value exposure, and compatible temporal change. We show that current action risk is unidentified without restricting unobserved label evolution. Under representative randomized audits, label-independent evidence windows, and a prespecified condition linking historical and current action risk, we derive simultaneous finite-sample control of unsafe authorization. Chronological evaluations with simulated audits on IEEE-CIS, ULB-Worldline, and Elliptic++ yield zero-drift automation rates of 84.4\%, 67.4\%, and 81.3\%, with total review workloads of 24.1\%, 46.0\%, and 43.1\%. The experiments reveal an audit-capacity trade-off: sparse auditing delays authorization, whereas intensive auditing eventually increases workload. A separately specified BAF stress test further indicates that fallback thresholds must reflect candidate-specific evidence rather than a common fraction of the risk limit. These findings identify audit freshness and analyst capacity as joint design considerations for fraud decision support.
\end{abstract}


\end{frontmatter}

\section{Introduction}

Financial institutions rarely translate a fraud score into a decision through a single threshold. Low-scoring events may be approved, ambiguous cases sent to an analyst, and high-scoring events blocked. The outcomes used to evaluate those actions often arrive later through investigations, disputes, chargebacks, or law-enforcement processes. Operational decisions also affect which outcomes become observable. Thus two periods with similar score distributions may warrant different policies if the available evidence is sparse or stale.

Most public fraud studies focus on discrimination or detection metrics. Research on realistic fraud evaluation, delayed labels, and selective labels shows why this view can misrepresent deployment performance \citep{dalpozzolo2018realistic,grzenda2020delayed,botacin2025delays,lakkaraju2017selective}. A high-ranking model may still provide too few audited observations to support automatic approval, while a permissive threshold may overwhelm investigators or concentrate losses in high-value transactions. Transaction value can also change the preferred fraud decision \citep{hoppner2022cost}. The deployment problem is therefore not only how to rank events, but also when the available evidence justifies acting on that ranking.

We examine this problem through three research questions:
\begin{enumerate}[label=\textbf{RQ\arabic*:},leftmargin=*]
    \item Under what information and temporal-stability conditions can delayed, selectively acquired audits certify the current risk of an automated action?
    \item How do audit rate, label delay, evidence age, and the action-risk limit jointly determine certifiable automation and total human workload?
    \item Which conclusions persist across financial-crime domains and frozen scorers, and when do count-risk and transaction-value assessments disagree?
\end{enumerate}
Together, these questions require separating predictive confidence from authority to automate while treating evidence age, diagnostic audits, manual decisions, non-authorization, and count/value exposure jointly.

We develop freshness-constrained audit capacity (\fcac) for this purpose. Given a fixed scoring model, \fcac{} evaluates candidate approve/review/block policies against finite-sample action-risk, evidence-freshness, and review-workload constraints. Its output is an auditable record of the action regions supported by the current evidence; unsupported regions remain in manual review. Because the score supplies candidate regions rather than authorization, the same decision procedure can be used with different model families.

The paper makes three contributions.
\begin{enumerate}[leftmargin=*]
    \item We formulate fraud automation as a decision-control problem in which predictive ranking and authority to act are separate. This formulation identifies evidence freshness, shared review capacity, non-authorization, and count/value exposure as linked design requirements.
    \item We develop and analyze a policy evaluator that joins mature randomized-audit evidence, a prespecified temporal allowance, asymmetric action-risk limits, and one workload ledger. An impossibility result establishes the information boundary for current-risk authorization; a conditional finite-sample result controls unsafe authorization over the candidate grid; and a candidate-specific frontier reports when no action region remains feasible.
    \item We evaluate the framework in chronological studies across three financial-crime domains, matched policy comparisons, workload and drift stresses, scorer sensitivity, and a separately specified BAF test. The results show how the same audit capacity both produces authorization evidence and consumes the analyst resource that automation is intended to conserve.
\end{enumerate}

\begin{figure}[!t]
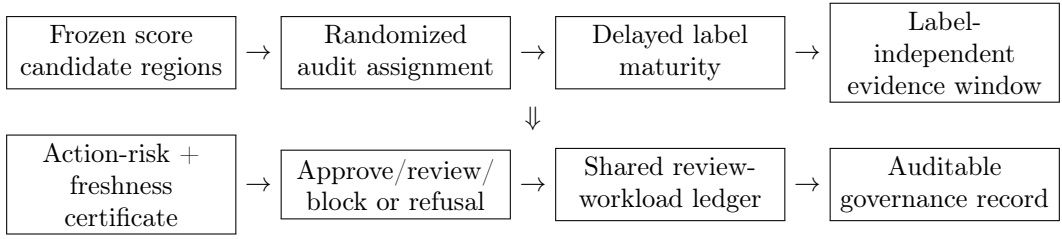

\centering
\footnotesize
\begin{tabular}{@{}c@{\ $\rightarrow$\ }c@{\ $\rightarrow$\ }c@{\ $\rightarrow$\ }c@{}}
\fbox{\parbox{0.17\textwidth}{\centering Frozen score\\candidate regions}} &
\fbox{\parbox{0.17\textwidth}{\centering Randomized\\audit assignment}} &
\fbox{\parbox{0.17\textwidth}{\centering Delayed label\\maturity}} &
\fbox{\parbox{0.17\textwidth}{\centering Label-independent\\evidence window}} \\
\multicolumn{4}{c}{\vspace{2pt}$\Downarrow$}\\[-2pt]
\fbox{\parbox{0.17\textwidth}{\centering Action-risk +\\freshness certificate}} &
\fbox{\parbox{0.17\textwidth}{\centering Approve/review/\\block or refusal}} &
\fbox{\parbox{0.17\textwidth}{\centering Shared review-\\workload ledger}} &
\fbox{\parbox{0.17\textwidth}{\centering Auditable\\governance record}}
\end{tabular}
\caption{\fcac{} decision flow. Risk owners specify action limits, confidence, and a temporal-stability allowance; operations provide score bands, audit rate, label delay, and workload limits. The output records authorized regions, evidence age, review demand, and the candidate-specific feasibility frontier.}
\label{fig:fcac-artifact}
\end{figure}

\fcac{} uses finite-sample risk control within a wider decision procedure. Adjacent work develops risk-controlling sets, anytime-valid labeling, online conformal methods under shift, and per-threshold certificates \citep{bates2021rcps,xu2024active,bao2025cap,gibbs2024arbitrary,khosravi2026selective}. \fcac{} instead asks whether asymmetric fraud actions may be delegated under delayed labels and shared review capacity, and reports the statistical decision with workload and a stability frontier.

\section{Related work}

\subsection{Fraud prediction and operational review}

Fraud research treats class imbalance, distributional change, and verification latency as related operational constraints \citep{dalpozzolo2018realistic}. Investigator decisions also create selective labels and an exploration--exploitation problem over verification \citep{carcillo2018streaming}. Hybrid scores, interpretable data engineering, and cost-sensitive learning improve detection, investigator support, or transaction-specific loss \citep{carcillo2021combining,baesens2021data,hoppner2022cost}; recent DSS research similarly combines domain representations with cost-sensitive evaluation \citep{hajek2026topic}. These approaches do not establish whether postdeployment labels support automation.

The Elliptic benchmarks support temporal blockchain-AML modeling \citep{weber2019elliptic,elmougy2023ellipticpp}, and ULB has been used to study imbalance and calibration \citep{dalpozzolo2015calibrating}. DSS research also links ex-ante fraud scores to downstream operational efficiency \citep{duan2024information}. Nanduri et al.'s production architecture is the closest precedent for the three-action setting \citep{nanduri2020microsoft}. We build on that operational view by studying the amount and recency of evidence needed to authorize each automated action.

\subsection{Selective prediction and risk control}

RCPS and conformal risk control provide finite-sample control of specified losses \citep{bates2021rcps,angelopoulos2024crc}. A-RCPS adds adaptive label queries and anytime validity \citep{xu2024active}; CAP controls online selective coverage statements \citep{bao2025cap}; and online conformal methods address arbitrary shift when the required structure is available \citep{gibbs2024arbitrary,prinster2024validity}. Recent work adds growing-sample anytime risk control and per-threshold selective acting \citep{hultberg2026anytime,khosravi2026selective}. In fraud DSS, DISCO combines deep metric learning with conformal false-negative-risk control and evaluates classification and operational efficiency \citep{zhu2026disco}.

Selective classification uses confidence or novelty to reject out-of-distribution inputs and is commonly evaluated through risk--coverage curves \citep{xia2024softmax}. These methods address label budgets, online validity, threshold-specific evidence, and abstention. \fcac{} additionally treats delayed label maturity, audit workload, and action-asymmetric errors; current authorization still requires an assumption connecting past audited outcomes to current action risk.

Table~\ref{tab:method-boundary} compares the decision objects and operational outputs of the closest risk-control approaches.

\begin{table}[htbp]
\centering
\caption{Decision objects and operational outputs of adjacent risk-control approaches.}
\label{tab:method-boundary}
\scriptsize
\begin{tabularx}{\textwidth}{p{0.15\textwidth}p{0.20\textwidth}p{0.19\textwidth}X}
\toprule
Approach & Primary supported object & Delayed/selective feedback & Relation to review operations \\
\midrule
RCPS / CRC \citep{bates2021rcps,angelopoulos2024crc} & Prediction sets or monotone loss control & Fixed calibration is the basic setting & Does not model the review ledger \\
A-RCPS \citep{xu2024active} & Anytime-valid risk control with adaptive label queries & Directly supported & Controls a label budget, but not a shared action-review ledger \\
CAP \citep{bao2025cap} & Online selective conformal coverage statements & Online selection and dynamic extensions & Does not model review operations \\
Anytime threshold control \citep{hultberg2026anytime,khosravi2026selective} & Growing-sample risk control or per-threshold selective acting & Sequential evidence supported & Does not integrate evidence age with review workload \\
DISCO \citep{zhu2026disco} & FNR-controlled fraud detection after representation learning & Rolling-window calibration & Operational efficiency without a diagnostic-audit workload \\
Static development CP & Frozen action threshold from development audits & No current-evidence update & Fixed policy plus matched monitoring accounting \\
\fcac{} & Conditional approve/block action-risk authorization & Explicit maturity delay and declared evidence age & Shared review workload, asymmetric actions, and explicit refusal \\
\bottomrule
\end{tabularx}
\end{table}

\subsection{Delayed and selectively observed labels}

Feedback timing changes model evaluation, and an instantaneous-label assumption can understate exposure when analysis queues are finite \citep{grzenda2020delayed,botacin2025delays}. Past decisions also affect which outcomes become observed \citep{lakkaraju2017selective}. Dynamic-budget active learning and adaptive financial policies use incoming information to respond to drift \citep{aguiar2024dynamic,wang2026financial}. \fcac{} addresses the complementary question of what the randomized audits that have already matured can support at the current decision time.

\subsection{Human--AI allocation}

DSS research has long examined how work should be divided between people and computational systems \citep{jones2002division}. Recent human--AI and design-science studies emphasize decision authority, interaction, transparency, and institutional constraints \citep{storey2024design,zolbanin2026transparent}. Learning-to-defer routes difficult cases to experts \citep{mozannar2020defer}, while cost-sensitive fraud methods optimize losses and thresholds \citep{hoppner2022cost}. In \fcac{}, allocation depends on both the score region and the evidence supporting automation; unsupported cases remain with analysts.

\section{Decision setting}

\subsection{Decision roles and artifact outputs}

\fcac{} is intended for a fraud-operations manager who must determine how a scoring model should be used in the decision process. Before deployment, the manager evaluates candidate policies against action-risk limits, confidence, audit rate, feedback delay, and a review-workload limit. The evaluator uses fixed score bands and the mature randomized-audit record. Temporal stability enters as a policy assumption and is not inferred from changes in the unlabeled score distribution.

For each decision period, the framework records which approve and block regions may be automated, which events remain in review, how many diagnostic audits add to the workload, and why a candidate region was not authorized. It also reports value exposure, cost sensitivity, and the maximum temporal-change rate compatible with at least one candidate. These outputs allow the manager to compare a proposed policy with the organization's operating limits.

The central resource coupling is that an audit of an otherwise automated event can enlarge the evidence base for future authorization while consuming one unit of current review capacity. Increasing the audit rate may therefore expand the supported action regions and reduce the manual region, but it also adds audit demand. Label delay postpones the evidence benefit while the workload cost remains immediate. \fcac{} reports both effects against the same capacity limit.

\subsection{Events, scores, and actions}

At native time period $t$, event $i$ has covariates $X_i$, transaction value $V_i\geq 0$, a frozen fraud score $S_i\in[0,1]$, and binary label $Y_i\in\{0,1\}$, where one denotes fraud. Two score thresholds define three actions:
\begin{equation}
a_i(\lambda_L,\lambda_H)=
\begin{cases}
\mathrm{approve}, & S_i\leq\lambda_L,\\
\mathrm{review}, & \lambda_L<S_i<\lambda_H,\\
\mathrm{block}, & S_i\geq\lambda_H.
\end{cases}
\end{equation}
The approve error is $E_i^A=Y_i$ and the block error is $E_i^B=1-Y_i$. The action-conditional count risks are
\begin{equation}
\risk_t^A(\lambda_L)=\Pr_t(Y=1\mid S\leq\lambda_L),\qquad
\risk_t^B(\lambda_H)=\Pr_t(Y=0\mid S\geq\lambda_H).
\end{equation}
The organization specifies asymmetric limits $\alpha_A$ and $\alpha_B$.

\subsection{Diagnostic audits and delayed maturity}

Each auditable event is independently queried with known probability $q_i>0$. Under the main protocol $q_i=q$, producing a uniform Bernoulli diagnostic audit. A queried label from event time $s$ becomes available at $s+d$, where $d$ is measured in native periods. An event in the manual-decision region already consumes one review. A diagnostic audit consumes an additional review only when the event would otherwise have been automated. At time $t$, certificate construction can access only queried labels whose maturity time is no greater than $t$; all other labels remain hidden during training and certification.

\subsection{Count risk and value risk}

Certification initially constrains count risk. Offline evaluation additionally reports value-weighted approve risk
\begin{equation}
\risk_{V,t}^A=\frac{\sum_{i\in A_t}V_iY_i}{\sum_{i\in A_t}V_i},
\end{equation}
and the analogous legitimate-value fraction among blocked events. Values are reported in native units within each dataset. For cross-domain sensitivity analysis, values are capped at the pre-test 99th percentile, transformed by $\log(1+V)$, and divided by their pre-test mean. No test outcome is used to fit this transformation.

\section{Information boundary for current action risk}

Let $\mathcal O_t$ be the sigma-field generated by all covariates and frozen scores observed through $t$, all audit decisions, and all labels that have matured by $t$.

\paragraph{Proposition 1 (observational non-identifiability)}
Let a decision procedure measurable with respect to $\mathcal O_t$ automate a nonempty current action set with positive probability. If the regular conditional distribution of unmatured current labels given $\mathcal O_t$ is unrestricted, then for every $\alpha<1$ there exist two data-generating laws that agree on the distribution of $\mathcal O_t$ but under which current action risk is respectively zero and one on the automated set. Hence no $\mathcal O_t$-measurable nontrivial certificate can uniformly guarantee current action risk at most $\alpha$.

\paragraph{Proof}
Fix an observable law $Q$ under which a nonempty set is automated. Extend it to $P_0$ and $P_1$ by assigning conditional action-error probabilities zero and one, respectively, to unmatured events in that set, while holding all other variables fixed. Both laws induce $Q$ on $\mathcal O_t$ and hence the same decision, but their current action risks are zero and one. No certificate with $\alpha<1$ is valid under both. $\square$

This information-set result does not preclude validity when the joint law, likelihood ratios, or shift structure is known \citep{prinster2024validity}.

\section{Freshness-constrained audit capacity}

\subsection{Operational decision procedure}

At each decision period, the evaluator follows a prespecified seven-step procedure:
\begin{enumerate}[leftmargin=*]
    \item draw or retrieve constant-propensity diagnostic-audit indicators;
    \item expose only audited labels that have matured under the declared delay;
    \item retrieve each candidate's preselected evidence window and compute its realized audit count, error count, and mean label age;
    \item add its KL index and temporal allowance, certifying it only when the sum does not exceed the action-risk limit;
    \item select the largest certified approve threshold and smallest certified block threshold; when a certified set is empty, disable automatic approval or automatic blocking, respectively;
    \item if the selected thresholds overlap, retain the certified approve region and disable automatic blocking for that period; and
    \item route all remaining events to manual decision and charge diagnostic audits of otherwise automated events to the same review ledger.
\end{enumerate}
The first four steps evaluate the evidence. The remaining steps convert supported candidates into a nonoverlapping approve/review/block policy. The overlap rule returns the entire would-be block region to review rather than trimming it into a new, uncertified subset. The resulting record gives the manager the action, evidence age, workload, and reason for non-authorization.

\subsection{Conditions for an operational certificate}

The authorization result is separate from the subsequent workload evaluation and relies on the following conditions:
\begin{enumerate}[label=A\arabic*,leftmargin=*]
    \item the score model and finite threshold grid are fixed before each held-out \fcac{} evaluation run;
    \item the baseline design sigma-field $\mathcal G_{tja}$ contains fixed covariates and scores, event times, candidate membership, randomized audit indicators, and the preselected evidence window, but no audit-error labels; conditional on pre-audit covariates and design information, the sigma-field generated by the entire randomized audit-selection vector is independent of the full vector of unobserved action-error labels, and audits have constant positive propensity within each candidate action region;
    \item with $\mathcal F_{i-1}=\mathcal G_{tja}\vee\sigma(Z_1,\ldots,Z_{i-1})$, audit errors form an adapted Bernoulli sequence with predictable conditional means $p_i=\Pr(Z_i=1\mid\mathcal F_{i-1})$ that may change after earlier outcomes;
    \item for each nonempty preselected evidence window, $\risk_t(a,j)\leq n^{-1}\sum_i p_i(a,j)+L_a\overline{\operatorname{age}}_{tja}$ almost surely under the data-generating law for a predeclared temporal-transport allowance rate $L_a$; the stronger audit-by-audit Lipschitz condition implies this window-average condition but is not required;
    \item the evidence window is selected without inspecting error labels; and
    \item confidence is allocated simultaneously across test times, thresholds, and actions.
\end{enumerate}
Nonuniform endogenous auditing requires a weighted confidence sequence. Error-label-adaptive selection among $M$ evidence windows would require an additional factor $M$ in the confidence allocation.

A4 is the temporal-transport condition that connects mature audit evidence to current action risk. Current score movements do not identify this condition. The organization therefore specifies $L_a$ before inspecting the audit errors, using a policy limit, historical evidence, or a stress value appropriate to the application. Proposition~1 shows why some restriction of this kind is required for current-risk authorization. A4 concerns the average predictable risk in the selected evidence window; an audit-by-audit Lipschitz condition is sufficient but stronger than the condition used in the proof.

\subsection{Fixed-limit KL certification under adaptive risks}

Order the mature audits in their preselected window and let $\mathcal F_{i-1}$ denote the history before error $Z_i$ is revealed. Its predictable conditional risk is $p_i=\Pr(Z_i=1\mid\mathcal F_{i-1})$, which may depend on earlier errors. For any fixed null risk $r$ and $q<r$, a Bernoulli test-martingale argument gives
\begin{equation}
\Pr\left(K/n\leq q,\ n^{-1}\sum_i p_i\geq r\right)
\leq\exp[-n\,\mathrm{kl}(q\Vert r)].
\end{equation}
This fixed-limit test follows from Markov's inequality applied to the nonnegative martingale $\prod_i\exp(\lambda Z_i)/(1-p_i+p_i\exp\lambda)$, optimized over $\lambda<0$ \citep{howard2021confidence}. Define $U^{\mathrm{KL}}(k,n,\delta)$ as the largest $p\geq k/n$ satisfying
\begin{equation}
n\,\mathrm{kl}(q\Vert p)=\log(1/\delta).
\end{equation}
For fixed realized $n$, null $r$, and allocation $\eta$, let
\begin{equation}
c(n,r,\eta)=\max\left\{k\in\{0,\ldots,n\}:k<nr,\ n\,\mathrm{kl}(k/n\Vert r)\geq\log(1/\eta)\right\},
\end{equation}
with $c=-1$ for an empty set. Monotonicity gives $U^{\mathrm{KL}}(K,n,\eta)\leq r$ exactly when $K\leq c$; applying the inequality at the fixed boundary $c/n$ controls the complete rejection event.
The study controls simultaneous threshold and action search through fixed action-specific grids and online alpha spending:
\begin{equation}
\delta_{tja}=\frac{6\delta}{\pi^2t^2J_a\,2},
\end{equation}
where $J_a$ is the number of retained candidates for action $a$ and there are two automated actions. These terms sum to at most $\delta$ over all times, thresholds, and actions.

\subsection{Current-risk lift under temporal stability}

\paragraph{Proposition 2 (simultaneous control of unsafe certification)}
Let $U^{\mathrm{KL}}_{tja}$ be the KL index computed from realized mature audits using $\delta_{tja}$. Under A1--A6, with A4 holding almost surely for every nonempty evidence window and zero-count windows never certified, the probability that \fcac{} certifies any tested time, threshold, or action for which $\risk_t(a,j)>\alpha_a$ is at most $\delta$. Each action region retained in the final policy is one of these certified candidates. When the selected thresholds overlap, the block action is disabled rather than truncated. Hence, with probability at least $1-\delta$, every automated region in the implemented policy satisfies its declared current-risk limit.

\paragraph{Proof}
Condition on $\mathcal G_{tja}$. By A2, conditioning on the whole randomized audit design does not reveal the hidden error vector. The audit count, label ages, and label-independent window are fixed, and the error process remains adapted as in A3. If the count is zero, the rule cannot certify. Otherwise the realized drift allowance $\rho=L_a\overline{\operatorname{age}}_{tja}$ is fixed. If $\rho\geq\alpha_a$, the rule again cannot certify. Otherwise set $r=\alpha_a-\rho$. By A4, $\risk_t(a,j)>\alpha_a$ implies $n^{-1}\sum_i p_i>r$. Certification is equivalent to $K\leq c(n,r,\delta_{tja})$. The fixed-limit martingale inequality at the nonrandom boundary $q=c/n$ bounds its joint occurrence with $n^{-1}\sum_i p_i\geq r$ by $\delta_{tja}$. Averaging this conditional bound over the random audit design preserves the allocation. A union bound and A6 control all candidates; A5 prevents an unaccounted error-label-adaptive window search. $\square$

\paragraph{Corollary 1 (declared-rate misspecification)}
Suppose A4 holds with rate $L_a^\star$, while the certificate is computed using a predeclared rate $\widehat L_a$. Under the remaining conditions of Proposition~2, with probability at least $1-\delta$, every certified candidate satisfies
\begin{equation}
\risk_t(a,j)\leq \alpha_a+
\bigl(L_a^\star-\widehat L_a\bigr)_+
\overline{\operatorname{age}}_{tja}.
\end{equation}
Thus overstatement of the required allowance preserves the risk limit but can reduce authorization, whereas understatement enlarges the guaranteed limit by at most the allowance shortfall times the candidate's mean evidence age.

\paragraph{Proof}
The simultaneous test event bounds the window mean by $\alpha_a-\widehat L_a\overline{\operatorname{age}}_{tja}$. Substitution into A4 at $L_a^\star$ proves the claim. $\square$

\paragraph{Corollary 2 (best-case zero-error capacity)}
When $k=0$, the KL index equals the familiar zero-error Clopper--Pearson expression $1-\delta^{1/n}$ \citep{clopper1934intervals}. If action limit is $\alpha$ and total drift allowance is $\rho<\alpha$, the exact minimum audit count is
\begin{equation}
n_0(\rho)=\left\lceil\frac{\log\delta_{\mathrm{eff}}}{\log[1-(\alpha-\rho)]}\right\rceil.
\end{equation}
If $\rho\geq\alpha$, no finite zero-error reference sample can certify current action risk under the declared policy.

\subsection{Evidence-window optimization}

Let $L$ be the maximum declared change in action risk per native period, $d$ the label delay, and $W$ the number of mature periods pooled. With approximately uniform action arrivals, mean label age is
\begin{equation}
\overline{\operatorname{age}}(W)=d+(W-1)/2.
\end{equation}
The window is selected before inspecting error labels. For its realized audit count $n_W$ and errors $k_W$, \fcac{} computes
\begin{equation}
B(W)=U_{\mathrm{KL}}(k_W,n_W,\delta_{\mathrm{eff}})+L\overline{\operatorname{age}}(W).
\end{equation}
A longer window reduces sampling uncertainty but increases staleness. The planning rule chooses $W^*$ from audit rate, action volume, delay, and drift budget; the operational rule evaluates the realized bound in that preselected window. Searching $M$ windows after seeing errors requires additional multiplicity correction.

Best-case phase diagrams use
\begin{equation}
n_{\mathrm{plan}}(W)=\lfloor qv_aW\rfloor,
\end{equation}
where $v_a$ is expected action-region arrivals per period, and set $k=0$. They identify infeasible configurations but do not authorize deployment.

\subsection{Governance choice of the temporal allowance}

An organization may set $L$ through a stress grid, a historical envelope, or the largest value compatible with a candidate action region. Operationally, $L$ is a risk-budget parameter rather than a point estimate of latent drift. A forecasting model may inform its setting, but the certificate requires a defensible upper temporal allowance. With unrestricted future change, worst-case analysis reduces to non-authorization by Proposition~1. Historical variation can inform the choice, while the future validity of A4 remains a maintained governance condition.

For fixed preselected windows and realized simultaneous statistical bounds, this reverse calculation is exact. Let $U_{tja}$ and $\bar g_{tja}$ denote the statistical upper bound and mean label age for candidate $j$, and let $\mathcal C_{ta}=\{j:N_{tja}>0,\ U_{tja}\leq\alpha_a\}$. Define
\begin{equation}
L^{\mathrm{crit}}_{ta}=\max_{j\in\mathcal C_{ta}}
\frac{\alpha_a-U_{tja}}{\bar g_{tja}},
\end{equation}
with no frontier when $\mathcal C_{ta}$ is empty and $L^{\mathrm{crit}}_{ta}=\infty$ if a statistically feasible candidate has zero age.

\paragraph{Proposition 3 (exact candidate feasibility frontier)}
For fixed candidate windows, counts, ages, and statistical bounds, at least one candidate in action family $a$ is certifiable at declared rate $L_a$ if and only if $\mathcal C_{ta}$ is nonempty and $L_a\leq L^{\mathrm{crit}}_{ta}$. Hence every candidate in that family is refused when $L_a>L^{\mathrm{crit}}_{ta}$.

\paragraph{Proof}
Candidate $j$ is certifiable exactly when $N_{tja}>0$ and $U_{tja}+L_a\bar g_{tja}\leq\alpha_a$. For positive age this is equivalent to $L_a\leq(\alpha_a-U_{tja})/\bar g_{tja}$; a feasible zero-age candidate is unaffected by $L_a$. Taking the maximum over statistically feasible candidates gives the claim. $\square$

The frontier reports the stability condition required to retain a candidate. Establishing A4 or selecting $L_a$ after examining the same audit errors requires evidence beyond this calculation.

\subsection{Human workload and break-even cost}

Human reviews equal manual-region events plus audits among automated events, with overlaps counted once. Relative to all-review, normalized utility is
\begin{equation}
\Delta U=c(N-H)-\eta V_{FA}+(1-\eta)V_{FB}-\gamma V_{LB},
\label{eq:utility}
\end{equation}
where $c$ is review cost, $H$ human reviews, $\eta$ reviewer sensitivity, $V_{FA}$ auto-approved fraud value, $V_{FB}$ auto-blocked fraud value, and $V_{LB}$ legitimate blocked value with friction $\gamma$. Setting $\Delta U=0$ gives the break-even cost; these are sensitivity parameters, not market prices.

\section{Experimental design}

\subsection{Datasets and chronological evaluation splits}

Table~\ref{tab:data} summarizes the three retrospective domains and the separate BAF stress stream. IEEE-CIS was released through the Vesta competition \citep{ieeecis2019dataset}, and ULB contains two days of European card transactions \citep{ulb2013dataset,dalpozzolo2015calibrating}. Elliptic++ retains the original Elliptic transaction labels and time steps and adds descriptors such as \texttt{total\_BTC} \citep{weber2019elliptic,elmougy2023ellipticpp}. We use only transaction labels and \texttt{total\_BTC}; actor labels are excluded from training, certification, and evaluation. Transactions with unknown labels are also excluded, so the reported Elliptic++ results apply only to the publicly labeled subset.

BAF Base contains one million privacy-preserving synthetic account applications generated from an anonymized source \citep{jesus2022turning}. We use three warm-up, two development, and three stress months under a protocol fixed before row-level inspection. Because BAF provides neither transaction values nor feedback timestamps, we assign unit values and simulate a one-month delay. The dataset is used only for the count-risk stress test.

\begin{table}[htbp]
\centering
\caption{Locked temporal data protocol. Counts in parentheses are fraud labels.}
\label{tab:data}
\small
\begin{tabularx}{\textwidth}{lXrrr}
\toprule
Dataset & Domain / native period & Warm-up & Development & Held-out test \\
\midrule
IEEE-CIS & E-commerce payment / relative day & 172,124 (4,806) & 84,291 (3,608) & 334,125 (12,249) \\
ULB-Worldline & Card transaction / hour & 47,401 (146) & 97,385 (135) & 140,021 (211) \\
Elliptic++ & Blockchain AML / time step & 24,200 (2,337) & 9,241 (1,380) & 13,123 (828) \\
BAF Base & Synthetic account opening / month & 397,039 (3,896) & 278,627 (2,844) & 324,334 (4,289) \\
\bottomrule
\end{tabularx}
\end{table}

To prevent temporal leakage, we partition each dataset chronologically into the warm-up, development, and held-out test periods shown in Table~\ref{tab:data}. For IEEE-CIS, these periods are days 0--41, 42--69, and 70--181, respectively; for ULB, hours 0--11, 12--23, and 24--47; and for Elliptic++, time steps 0--25, 26--37, and 38--49. As noted above, only transactions with known labels are retained in each Elliptic++ period.

Here ``held out'' denotes a temporal test block whose outcomes were excluded from fitting the scorer, thresholds, evidence windows, and transformations; no unmatured, unaudited label entered a certificate. Because these public datasets also informed earlier research development, the study is a controlled retrospective temporal evaluation rather than an independent confirmation.

\subsection{Scorer, thresholds, and audit grid}

The common scorer is XGBoost \citep{chen2016xgboost} with 50 trees fitted once on the warm-up data. We treat the scorer as an input to the decision framework. For each action, up to 40 quantile thresholds are set from development scores, with duplicate thresholds removed. This yields 40 approve and 40 block candidates for IEEE, 34 and 40 for ULB, and 39 and 16 for Elliptic++. Every sensitivity scenario reuses the same cached scores, and the simultaneous confidence budget uses the retained action-specific count $J_a$.

The experimental grid uses audit rates $q\in\{0.05,0.10,0.20,0.30\}$, delays $d\in\{0,1,3,7\}$, and ten audit seeds. Approve limits are 2\% for IEEE and Elliptic++ and 0.1\% for ULB; the legitimate-block limit is 5\%. These values define illustrative operating policies; regulatory choices require application-specific calibration. Because the limits and native periods differ, the results do not support a ranking of datasets. The phase study also varies action coverage (25\%, 50\%, 75\%) and $L/\alpha\in\{0,0.5\%,1\%,2.5\%,5\%,10\%\}$. Proposition~2 applies to a policy specified before outcomes are observed, not to an outcome-driven choice from this grid.

The operational study fixes delay at three periods and audit rates of 10\%, 20\%, and 30\% for IEEE, ULB, and Elliptic++. Each evidence window is selected before evaluation from development scores, action volume, audit rate, delay, and the zero-error bound; certificates then use realized counts, errors, and ages. The protocol produces 8,880 decision-period evaluations across ten seeds and six drift allowances.

The separate 180-row BAF protocol uses 50 trees, 40 candidates per action, 10\% auditing, a one-month delay, 2\% approve and 5\% block limits, ten seeds, and six drift rates. Its prespecified qualitative endpoint requires automation to be nonincreasing and to reach all-review at 10\% drift for every seed. The practical checks require at least 10\% zero-drift automation and no more than 2\% fraud among auto-approved cases. The endpoint is evaluated without retuning.

An exploratory class-balanced logistic sensitivity analysis uses median imputation and standardization. It retains all splits, policies, and \fcac{} rules and adds a second 8,880-row evaluation.

A post-hoc zero-drift analysis adds 200 new audit seeds (29,600 rows) while holding all other inputs fixed. It measures variation due to randomized audit assignment rather than uncertainty about a future transaction population.

\subsection{Evaluation and validation}

The outcomes are certificate availability, restricted time to first certificate, automation and total human-review coverage, and count and normalized-value risk. Non-certification is coded one period beyond the evaluation horizon. We report means and empirical 2.5th and 97.5th percentiles for ten diagnostic-audit seeds and, post hoc, 200 new seeds. With the transaction stream and scorer fixed, these percentiles measure audit-assignment sensitivity rather than future-population uncertainty. Phase diagrams use all auditable post-warm-up history at the held-out endpoint. Exact-binomial experiments check certification probabilities and robust type-I error; a post-hoc stress sets the declared allowance to 0\%, 50\%, 100\%, or 150\% of the true allowance.

\section{Results}

\subsection{Audit rate and review workload}

\begin{figure}[htbp]
\centering
\includegraphics[width=\textwidth]{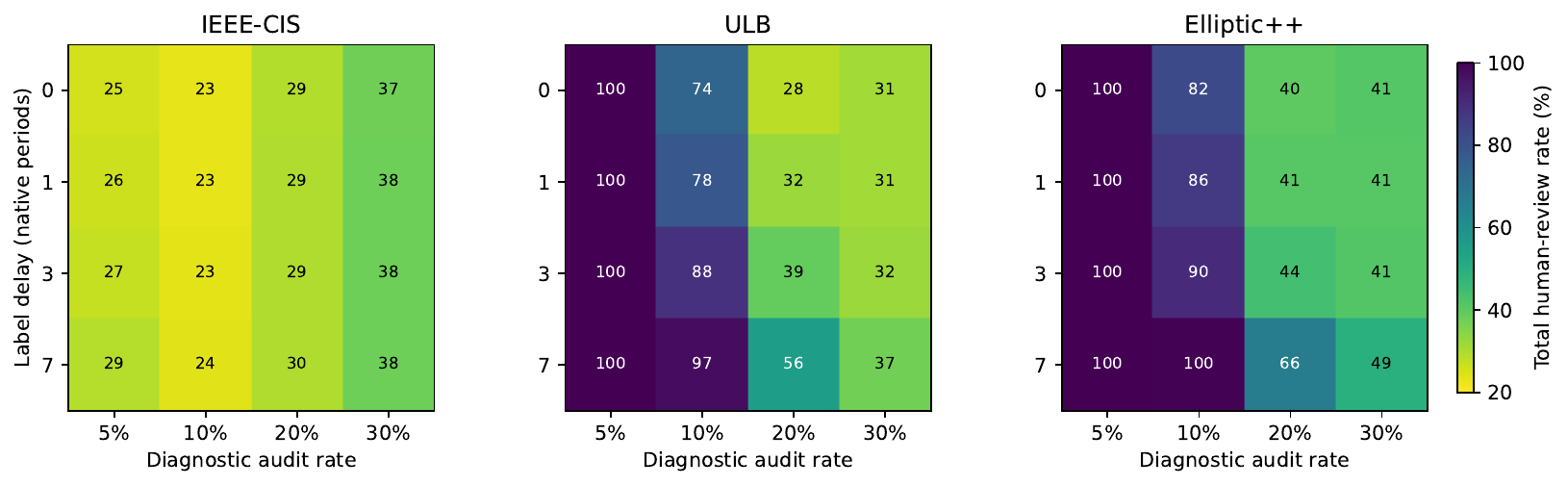}
\caption{Total human-review rate across audit-rate and delay scenarios. Values include the manual-decision region and diagnostic audits of otherwise automated events.}
\label{fig:heatmap}
\end{figure}

Figure~\ref{fig:heatmap} reports total human workload, including diagnostic audits among automated events. At delay three, IEEE audit rates of 5\%, 10\%, 20\%, and 30\% yield human-review rates of 27.0\%, 23.2\%, 29.5\%, and 37.6\%. The 10\% policy is an interior workload optimum: 5\% supplies too little evidence and leaves a larger manual region, whereas higher rates increasingly consume analysts through diagnostic checks. At zero delay, the minimum observed workload occurs at 20\% for ULB (28.4\%) and Elliptic++ (39.8\%). The optimum is therefore domain- and delay-dependent.

\begin{figure}[htbp]
\centering
\includegraphics[width=\textwidth]{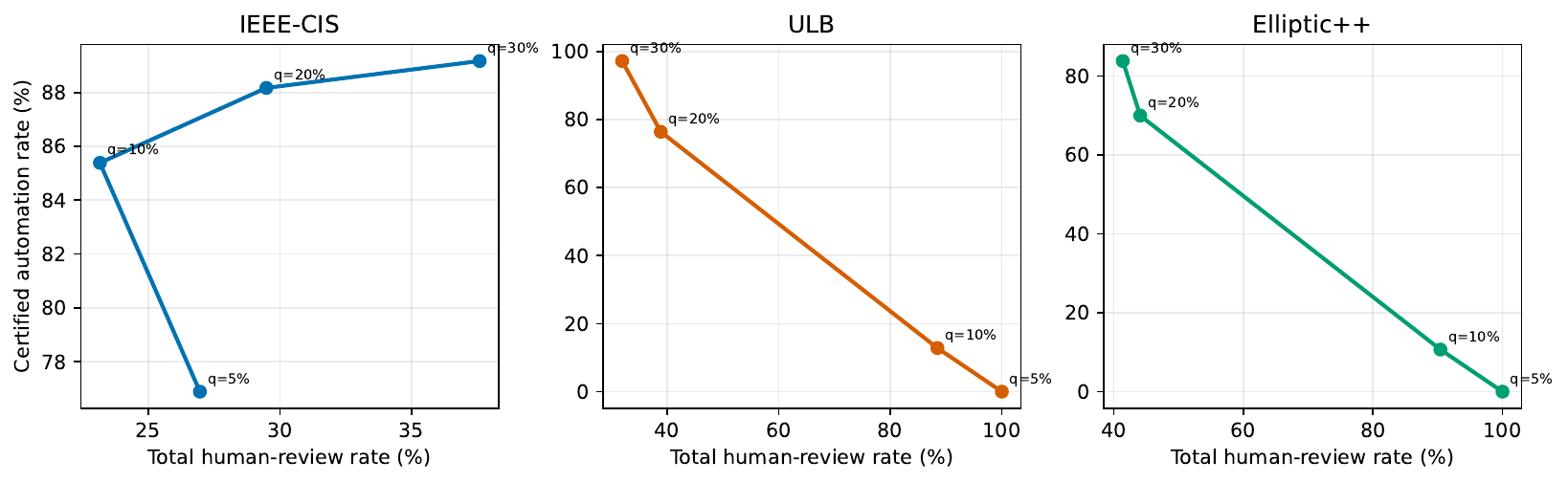}
\caption{Certified automation versus total human-review workload at label delay three.}
\label{fig:tradeoff}
\end{figure}

Figure~\ref{fig:tradeoff} shows the operational frontier at delay three. For ULB and Elliptic++, low audit rates produce near-all-review because evidence is insufficient. For IEEE, too much diagnostic auditing dominates the workload once certification is already broadly available.

\subsection{Effect of label delay on automation}

At a 10\% ULB audit rate, certificate availability falls from 25.4\% at zero delay to 21.7\%, 12.1\%, and 2.9\% for delays 1, 3, and 7. Restricted time to first certificate rises from 16.4 to 17.2, 22.1, and 24.3 within a 24-period test horizon. At a 20\% Elliptic++ audit rate, automation falls from 75.3\% at zero delay to 42.4\% at delay seven.

\subsection{Matched authorization-policy comparison}

Table~\ref{tab:baseline} compares all-review, a static development CP policy, a matched delayed-anytime policy, and zero-drift \fcac{}. The policies share the transaction stream, cached scores and threshold grid, audit rate, label delay, seeds, action limits, and family-wise confidence budget; each sequential method retains its own time-uniform boundary. Static CP sets thresholds from development audits and is charged the same monitoring workload. The delayed-anytime comparator uses mature current audits without a temporal allowance. The comparison therefore aligns the operating conditions while preserving the distinct authorization rules.

\begin{table}[htbp]
\centering
\caption{Matched zero-drift policy comparison. Risks are pooled realized approve/block errors (A/B); period exceedance is descriptive and is not the event controlled by Proposition~2. First authorization is the restricted mean period, with non-authorization coded as horizon plus one.}
\label{tab:baseline}
\scriptsize
\begin{tabular}{llrrrrr}
\toprule
Dataset & Policy & Auto. & Review & First & Risk A/B & \shortstack{Action-period\\exceed.} \\
\midrule
\multirow{4}{*}{IEEE-CIS} & All review & 0.0\% & 100.0\% & 113.0 & -- & -- \\
& Static development CP & 81.9\% & 26.3\% & 1.0 & 1.303\%/-- & 3.9\% \\
& Matched delayed-anytime & 85.4\% & 23.2\% & 1.0 & 1.427\%/-- & 8.2\% \\
& \fcac{} ($L=0$) & 84.4\% & 24.1\% & 1.0 & 1.387\%/-- & 6.2\% \\
\addlinespace
\multirow{4}{*}{ULB} & All review & 0.0\% & 100.0\% & 25.0 & -- & -- \\
& Static development CP & 88.6\% & 29.1\% & 3.4 & 0.0181\%/-- & 0.0\% \\
& Matched delayed-anytime & 76.4\% & 38.9\% & 3.8 & 0.0161\%/-- & 0.0\% \\
& \fcac{} ($L=0$) & 67.4\% & 46.0\% & 6.8 & 0.0138\%/-- & 0.6\% \\
\addlinespace
\multirow{4}{*}{Elliptic++} & All review & 0.0\% & 100.0\% & 13.0 & -- & -- \\
& Static development CP & 86.6\% & 39.4\% & 1.0 & 1.987\%/0.637\% & 20.8\% \\
& Matched delayed-anytime & 83.8\% & 41.4\% & 1.0 & 1.958\%/0.592\% & 20.0\% \\
& \fcac{} ($L=0$) & 81.3\% & 43.1\% & 1.0 & 1.931\%/0.537\% & 19.6\% \\
\bottomrule
\end{tabular}
\end{table}

No policy dominates on efficiency in all three datasets. Relative to delayed-anytime, \fcac{} uses 0.9, 7.2, and 1.7 percentage points more review workload in IEEE, ULB, and Elliptic++, respectively. This difference is the operating cost of charging evidence age against the risk limit. The period-exceedance frequency in Table~\ref{tab:baseline} is descriptive and differs from the simultaneous unsafe-authorization event controlled by Proposition~2. The raw value/count-risk ratios are 1.21, 3.17, and 0.043 for IEEE, ULB, and Elliptic++; the normalized ratios are 1.12 and 0.20 for ULB and Elliptic++. Count-risk control therefore leaves economically relevant variation in value exposure. Under the prespecified cost sensitivity, the mean normalized break-even review costs are 0.0150, 0.000172, and 0.00670. They are scenario-specific sensitivity measures; market-price interpretation would require external cost data.

\subsection{Automation under temporal-stability allowances}

\begin{figure}[htbp]
\centering
\includegraphics[width=0.92\textwidth]{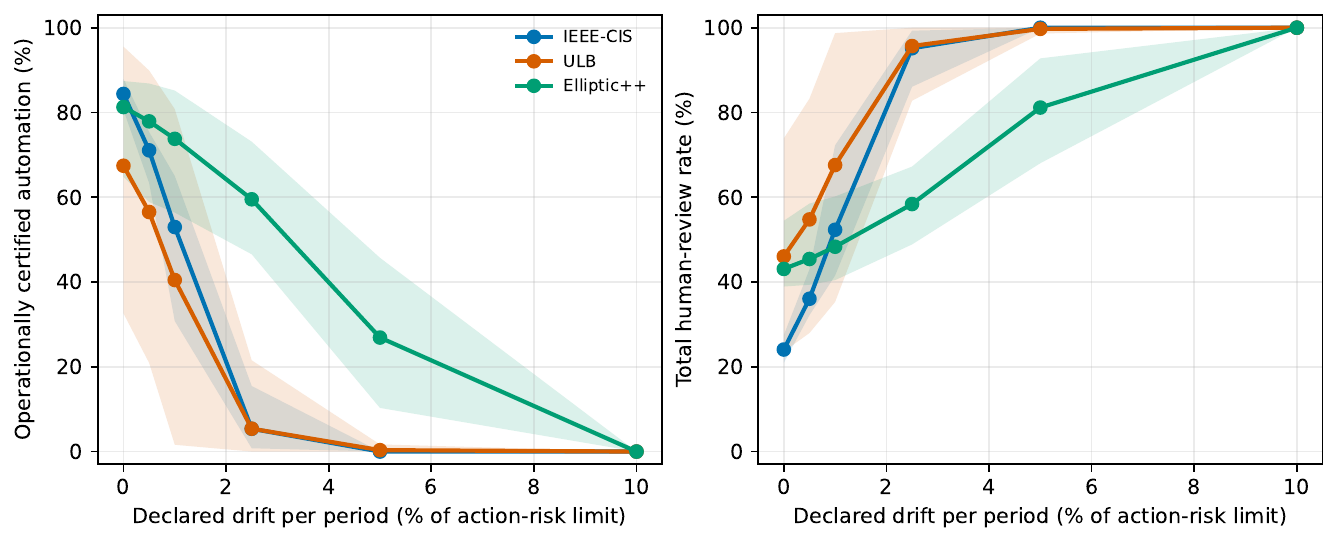}
\caption{Operational \fcac{} response using realized audit counts, errors, and label ages. Lines are audit-seed means; shading is the empirical 2.5th--97.5th percentile. The horizontal axis declares $L/\alpha$ per native period only; candidate decisions use $L\bar g/\alpha$, so cross-domain slopes are not comparable.}
\label{fig:empirical-fcac}
\end{figure}

Figure~\ref{fig:empirical-fcac} reports the complete operational calculation using realized audit counts, errors, and label ages. At zero temporal allowance, mean certified automation is 84.4\% for IEEE, 67.4\% for ULB, and 81.3\% for Elliptic++. The corresponding empirical 2.5th--97.5th percentile ranges across audit seeds are [80.6\%, 87.9\%], [32.6\%, 95.5\%], and [64.8\%, 87.4\%]. ULB is especially sensitive to which events are audited, whereas IEEE varies much less. After diagnostic audits of automated events are included, the mean human-review rates are 24.1\%, 46.0\%, and 43.1\%. Mean count risks among auto-approved cases are 1.384\%, 0.0134\%, and 1.931\%, below the respective limits of 2\%, 0.1\%, and 2\%. These realized outcomes describe the evaluated streams; the authorization guarantee remains the conditional statement in Proposition~2.

In the post-hoc 200-seed zero-drift study, mean automation is 83.3\%, 68.9\%, and 79.9\%; the largest difference from the primary ten-seed means is 1.47 percentage points. The wider ranges, particularly for ULB, provide a more detailed view of randomized-audit sensitivity.

A post-hoc common-limit analysis confirms that the risk limit is a major determinant of capacity. At a 0.1\% approve limit, automation is 0\%, 67.4\%, and 3.4\% for IEEE, ULB, and Elliptic++, respectively. Each limit defines a separate policy and was evaluated as such.

Increasing the temporal-change rate reduces automation monotonically in the three retrospective streams (Table~\ref{tab:empirical-fcac}). Since the native periods differ, the comparable operating quantity is the dimensionless cumulative staleness share $L\bar g/\alpha$, not $L/\alpha$ by itself. At the 2.5\%-per-period grid point, staleness consumes an average of 37.3\%, 27.9\%, and 22.1\% of the approve-risk limit for the selected IEEE, ULB, and Elliptic++ regions, while automation falls to 5.4\%, 5.4\%, and 59.5\%. At 10\%, none of the three retrospective streams authorizes an automated action, although the BAF result below shows that this endpoint is dataset-specific. Only Elliptic++ produces block certificates.

\begin{table}[htbp]
\centering
\caption{Mean authorized automation across ten seeds. The first column is a declared rate per native period; cross-domain interpretation uses candidate-specific $L\bar g/\alpha$.}
\label{tab:empirical-fcac}
\begin{tabular}{rrrrr}
\toprule
$L/\alpha$ per native period & IEEE-CIS & ULB & Elliptic++ & BAF Base \\
\midrule
0\% & 84.4\% & 67.4\% & 81.3\% & 99.3\% \\
0.5\% & 71.1\% & 56.5\% & 77.9\% & 99.3\% \\
1\% & 53.0\% & 40.5\% & 73.8\% & 99.3\% \\
2.5\% & 5.4\% & 5.4\% & 59.5\% & 99.3\% \\
5\% & 0\% & 0.3\% & 26.9\% & 99.3\% \\
10\% & 0\% & 0\% & 0\% & 99.2\% \\
\bottomrule
\end{tabular}
\end{table}

A controlled post-hoc stress holds the audit errors, KL rule, and confidence level fixed across three risk limits, true allowances of $0.25\alpha$ or $0.50\alpha$, and three audit-count levels; unsafe current risk is $1.05\alpha$. At declared-to-true allowance ratios of 0, 0.5, 1, and 1.5, the largest false-authorization rates are 99.96\%, 60.59\%, 4.06\%, and 0.045\%, respectively. Overstatement lowers mean safe-authorization power from 29.5\% under correct specification to 1.2\%. This asymmetry accords with Corollary~1 and leaves application-specific selection of $L$ as a governance decision.

\subsection{From certification to a decision record}

Table~\ref{tab:decision-record} illustrates how the statistical calculation becomes an operational record. The example is selected by evaluation order: the first held-out IEEE-CIS period for the first prespecified audit seed under the primary 10\% audit rate, three-period delay, and zero temporal allowance. The evidence supports automatic approval, while automatic blocking remains unavailable and the rest of the period is assigned to review.

\begin{table}[htbp]
\centering
\caption{Illustrative \fcac{} decision record for IEEE-CIS period 70, audit seed 7.}
\label{tab:decision-record}
\small
\begin{tabularx}{\textwidth}{p{0.20\textwidth}p{0.38\textwidth}X}
\toprule
Record element & Recorded value & Decision use \\
\midrule
Candidate action & Approve if $S\leq0.0373$; block unavailable & Defines the action that may be delegated \\
Mature evidence & 6,643 audits; 85 errors; mean age 15.21 days & Establishes the sample size, errors, and evidence freshness \\
Risk calculation & KL index 1.906\%; temporal allowance 0\%; approve limit 2\% & Authorizes the approve region \\
Period allocation & 2,777 auto-approved; 490 manual; 0 auto-blocked & Converts the certificate into an action policy \\
Review ledger & $490+308=798$ reviews (24.4\% of 3,267 events) & Combines manual decisions and diagnostic audits \\
Observed outcome & 41/2,777 approve errors (1.48\%) & Retrospective diagnostic unavailable at decision time \\
\bottomrule
\end{tabularx}
\end{table}

\subsection{Workload variability and capacity thresholds}

Let $C$ denote the review capacity per incoming event. At zero drift, mean workload is 24.1\%, 46.0\%, and 43.1\% for IEEE, ULB, and Elliptic++, but the seed-period q50/q95/q99 values are 23.2/32.4/38.0\%, 34.6/100/100\%, and 40.4/54.8/65.4\%. An illustrative 30\% limit is exceeded in 8.8\%, 55.8\%, and 99.2\% of seed-period observations; a 50\% limit is exceeded in 0\%, 40.8\%, and 10.8\%. A policy may therefore violate a prespecified workload quantile even when its mean workload appears feasible. The calculation counts reviews but does not model service times, analyst shifts, or queue dynamics.

\subsection{BAF stress-test outcome}

BAF passes the zero-drift practical checks but retains 99.21\% automation at the prespecified 10\% endpoint. All ten seeds therefore fail the required all-review condition. A separate implementation reproduces all 30 maximum-drift seed--period decisions. For the realized one-period evidence, the post-hoc dimensionless frontier $L^{\mathrm{crit}}\bar g/\alpha$ ranges from 71.1\% to 91.3\% of the risk budget. Thus a common value of $L/\alpha$ does not provide a universal fallback trigger; the stability allowance must be justified for the application and its evidence ages.

\subsection{Scorer sensitivity}

The exploratory scorer analysis reduces zero-drift automation from 84.4\%, 67.4\%, and 81.3\% under XGBoost to 35.6\%, 52.7\%, and 36.4\% under logistic scores. The authorization procedure can therefore be used with different model families, but the capacity it delivers depends on how the scorer partitions risk. Automation decreases monotonically over the drift grid for both scorers; BAF is not included in this post-hoc comparison.

\subsection{Best-case feasibility regions}

Table~\ref{tab:phase} aggregates audit rate, action coverage, and delay under the best-case planning assumption of zero errors. The strict 0.1\% ULB approve limit leaves little room for stale evidence: when per-period drift equals 5\% of the action-risk limit, only 2.1\% of the configurations remain feasible; at 10\%, none do. The table identifies combinations of risk, delay, stability, and audit capacity that cannot be reconciled even in the zero-error case. Actual authorization still depends on realized audit evidence.

\begin{table}[htbp]
\centering
\caption{Planning-only fraction of zero-error expected-audit capacity scenarios with at least one certifiable evidence window.}
\label{tab:phase}
\begin{tabular}{rrrr}
\toprule
Drift per period / risk limit & IEEE-CIS & ULB & Elliptic++ \\
\midrule
0\% & 100.0\% & 50.0\% & 75.0\% \\
0.5\% & 100.0\% & 50.0\% & 75.0\% \\
1\% & 100.0\% & 41.7\% & 75.0\% \\
2.5\% & 91.7\% & 20.8\% & 62.5\% \\
5\% & 66.7\% & 2.1\% & 35.4\% \\
10\% & 31.3\% & 0\% & 8.3\% \\
\bottomrule
\end{tabular}
\end{table}

A post-hoc development-only diagnostic uses the same randomized-audit design to construct simultaneous Clopper--Pearson intervals for each action region and period and to bound adjacent-period risk increases. No candidate falls below its own best-case critical frontier. For approve regions, the median historical envelopes are 7.94, 15.77, and 13.63 times the action limit in IEEE, ULB, and Elliptic++. These large values reflect missing audits and wide intervals rather than estimates of extreme underlying drift. The development histories are too sparse to calibrate a useful value of $L$, although they can inform stress scenarios.

\subsection{Validation checks}

The analytical power calculation agrees with exact-binomial simulation to within 0.44 percentage points. Across 27 heterogeneous and nine history-adaptive Bernoulli nulls, the largest false-certificate rate was 4.921\% at a nominal 5\% level. Exact enumeration of 60 adaptive designs and mixing over 1,782 count-conditional and 162 random-count cases found no violation of the allocated event bound. Fifty-one implementation tests cover the discrete boundary, critical-drift frontier, audit-count mixing, BAF schema guards, and realized label ages.

\section{Decision-support implications and transferable design principles}

The analysis reveals a coupling that is absent from score-only deployment rules. Diagnostic audits produce the evidence required for authorization, yet audits of automated events consume the same analyst capacity that automation is intended to release. Label delay weakens both sides of this relation by reducing the mature sample and increasing evidence age. Five design implications follow.

\paragraph{D1: Separate predictive confidence from authority to automate} A fixed score can rank candidate regions, but Proposition~1 shows that mature labels and current scores alone cannot provide a nontrivial guarantee for current automation. Authorization should therefore depend on representative, mature evidence in addition to the score.

\paragraph{D2: Treat evidence freshness and risk appetite as policy inputs} Tighter action-risk limits require disproportionately more audits from the relevant action region, while older evidence consumes more of the risk budget. The temporal allowance should be set before audit errors are inspected. The critical-drift frontier then shows whether the proposed stability assumption is compatible with any candidate region.

\paragraph{D3: Use one workload ledger for manual decisions and diagnostic audits} Review demand is not simply one minus the automation rate because some automated events must still be audited. The period-level analysis also shows that a mean workload can conceal severe peaks. Organizations should specify a workload quantile or breach tolerance while recognizing that event counts alone do not constitute a queueing or staffing model.

\paragraph{D4: Make non-authorization an informative output} When sampling uncertainty and staleness cannot both fit below the action-risk limit, the system returns the affected region to review and records the binding constraint. The BAF result shows that this decision should be based on candidate-specific evidence ages and bounds, not on a common drift-grid endpoint.

\paragraph{D5: Govern count and value exposure separately} An approve region that satisfies a count-risk limit may still concentrate high-value fraud, while a higher count risk can correspond to modest monetary exposure in another domain. The risk measure used for authorization should therefore be specified for each action. In the present implementation, value measures support this assessment but are not covered by the certificate.

\section{Limitations}

This study evaluates computational and statistical feasibility rather than organizational impact. The public datasets contain neither real diagnostic-audit assignments nor analyst decisions, so both label delay and randomized auditing are simulated according to prespecified protocols. The results consequently do not establish user adoption, changes in analyst behavior, realized savings, fairness, or causal effects in production. The three retrospective streams also informed earlier stages of the research and should be viewed as controlled temporal evaluations. Although the BAF protocol was specified before the data were acquired, BAF is privacy-preserving synthetic data, requires a simulated one-month delay and unit transaction values, and fails its qualitative endpoint.

The finite-sample guarantee depends on several substantive conditions: representative audits drawn with constant propensity, conditional independence between the complete randomized selection vector and hidden action errors, predictable conditional Bernoulli risks, evidence-window selection that does not use error labels, and a prespecified bound on average temporal change within the window. Marginal audit propensities and pairwise label independence do not suffice. The result allows conditional risks to adapt to past outcomes, but it does not cover endogenous auditing, arbitrary interference, or uncorrected searches over windows after their errors have been observed. The temporal-change rate remains a governance input because unlabeled score drift does not identify it. Corollary~1 quantifies conditional degradation when the declared rate differs from a valid rate, but neither it nor the post-hoc stress estimates that valid rate. Similarly, searching the reported policy grid using realized audit errors would require an additional selection correction, and the critical-drift frontier only describes compatibility with the chosen bound; it cannot verify the bound itself.

The current certificate controls count risk, not value risk or the error fraction in every realized batch. When the independently selected action thresholds overlap, the implementation prioritizes approval by disabling automatic blocking; alternative action priorities may produce different policies. \fcac{} also evaluates proposed audit rates instead of optimizing them. Workload quantiles count unit reviews without representing service times, queues, or differences among analysts. The phase diagrams use expected audit counts and describe planning feasibility only. Cross-domain comparisons are further limited by different native periods and monetary units, ULB's two-day horizon, and the restriction of Elliptic++ to publicly labeled transactions. Finally, the evaluator searches a finite policy grid and provides no online regret guarantee.

\section{Conclusion}

This study reframes fraud automation as a question of whether current evidence is adequate for a proposed action. With respect to RQ1, mature labels and current scores alone cannot provide a nontrivial guarantee when the evolution of unobserved outcomes is unrestricted; representative randomized audits, label-independent evidence windows, and a prespecified temporal-transport condition make conditional finite-sample control possible. For RQ2, the empirical results show that audit rate has a non-monotone relationship with workload and that label delay can remove otherwise feasible automation. For RQ3, capacity varies substantially across datasets and scoring models, and count-risk control does not determine value exposure. The BAF stress test further shows why fallback rules must be tied to candidate evidence rather than a common drift fraction. By governing evidence freshness and analyst capacity together, \fcac{} gives fraud-operations managers an auditable basis for delegating supported actions and retaining human review elsewhere.

\section*{Data and code availability}

The study uses public IEEE-CIS, ULB-Worldline, Elliptic++, and BAF Base data subject to their respective access terms; raw files and prepared caches are not redistributed. A verified reproduction package containing code, frozen configurations, checksums, derived results, and reproduction scripts is available from the corresponding author during editorial review and will be deposited in a public archival repository upon acceptance. No proprietary Binance or customer data are used.

\bibliographystyle{elsarticle-num}
\bibliography{references}

\end{document}